\documentclass[10pt,journal,compsoc]{IEEEtran}
\ifCLASSOPTIONcompsoc
  \usepackage[nocompress]{cite}
\else
  \usepackage{cite}
\fi
\ifCLASSINFOpdf
\else
\fi
\usepackage[dvipdfmx]{hyperref}
\usepackage{graphicx}
\usepackage{subfigure}
\usepackage{comment}

\begin{document}

%
\title{cvpaper.challenge in 2016: Futuristic Computer Vision through 1,600 Papers Survey}
%
%
%
%

\author{Hirokatsu~Kataoka,~Soma~Shirakabe,~Yun~He,~Shunya~Ueta,~Teppei~Suzuki,~Kaori~Abe,~Asako~Kanezaki,~Shin'ichiro Morita,~Toshiyuki~Yabe,~Yoshihiro~Kanehara,~Hiroya~Yatsuyanagi,~Shinya~Maruyama,~Ryosuke~Takasawa,~Masataka~Fuchida,~Yudai~Miyashita,~Kazushige~Okayasu,~Yuta~Matsuzaki
\IEEEcompsocitemizethanks{\IEEEcompsocthanksitem H. Kataoka and A. Kanezaki are with National Institute of Advanced Industrial Science and Technology (AIST).\protect\\
E-mail: see http://hirokatsukataoka.net/
}%
\IEEEcompsocitemizethanks{\IEEEcompsocthanksitem S. Shirakabe, S. Ueta and Y. He are with National Institute of Advanced Industrial Science and Technology (AIST) and University of Tsukuba.\protect\\
}%
\IEEEcompsocitemizethanks{\IEEEcompsocthanksitem T. Suzuki is with National Institute of Advanced Industrial Science and Technology (AIST) and Keio University.\protect\\
}%
\IEEEcompsocitemizethanks{\IEEEcompsocthanksitem K. Abe, S. Morita, Y. Matsuzaki, K. Okayasu, T. Yabe, Y. Kanehara, H. Yatsuyanagi, S. Maruyama, R. Takasawa, M. Fuchida, Y. Miyashita are with Tokyo Denki University.\protect\\
}%
\thanks{Manuscript received July 20, 2017; revised July 21, 2017.}}

%
%

\markboth{cvpaper.challenge in 2016, July~2017}%
{Kataoka \MakeLowercase{\textit{et al.}}: Bare Demo of IEEEtran.cls for Computer Society Journals}
%



\IEEEtitleabstractindextext{%
\begin{abstract}
The paper gives futuristic challenges disscussed in the cvpaper.challenge. In 2015 and 2016, we thoroughly study 1,600+ papers in several conferences/journals such as CVPR/ICCV/ECCV/NIPS/PAMI/IJCV.
\end{abstract}

}

\maketitle

\IEEEdisplaynontitleabstractindextext

%
\IEEEpeerreviewmaketitle


\section{Introduction}
\IEEEPARstart{I}{n} the last decade, the computer vision field has greatly developed as the results of revolutional techniques. Image representations like SIFT~\cite{LoweIJCV2004}, Haar-like~\cite{ViolaCVPR2001} and HOG~\cite{DalalCVPR2005} have made dramatically advances in the problems, both specified object recognition and image matching. More advanced algorithms achieved general object recognition (e.g. BoW~\cite{CsurkaECCVW2004}, SVM) and 3D reconstruction in a large space~\cite{AgarwalICCV2009}. However, the most influenced algorithm is to be a deep convolutional neural networks (DCNN) in 2012. The most significant CNN result was obtained by AlexNet in the ILSVRC2012~\cite{KrizhevskyNIPS2012}, which remains the image recognition leader with 1,000 classes. The effectiveness of transfer learning and parameter fine-tuning was shown with a pre-trained DCNN model~\cite{DonahueICML2014}. Recently, the DCNN framework is assigned in various problems such as stereo matching~\cite{ZbontarCVPR2015}, 3D recognition~\cite{SuICCV2015}, motion representation~\cite{SimonyanNIPS2014}, denoising~\cite{XieNIPS2012}, edge detection~\cite{BertasiusCVPR2015} and optical flow~\cite{WeinzaepfelICCV2013}. Undoubtedly, the current area of interest has been shifting a next step, such as temporal analysis, unsupervision/weak supervision, worldwide analysis, generative model and reconstruction.

However, the most important thing is \textit{to devise ideas to make a new problem}. The recent conferences such as CVPR~\cite{CVPR2017BNMW}, ACM MM~\cite{ACMMM2017BraveNewIdeas} and PRMU~\cite{PRMU201612} have organized a couple of workshops to find brave new ideas in the future. Especially in the PRMU, the community has thought futuristic technologies in computer vision since 2007. The PRMU decided 10 futuristic problems as grand challenges in 2009. They recognized image understanding including semantic segmentation and image captioning is the most difficult problem, however, the problem has solved by a recent DCNN. (e.g. semantic segmentation~\cite{LongCVPR2015,BadrinarayananarXiv2015} and image captioning~\cite{DonahueCVPR2015,JohnsonCVPR2016,YangCVPR2016}) Here, brave ideas are required in the futuristic computer vision techniques!

On one hand, the authors have promoted the project named \textit{cvpaper.challenge}. The \textit{cvpaper.challenge} is a joint project aimed at reading \& writing papers mainly in the field of computer vision and pattern recognition~\footnote{Further reading: \href{https://twitter.com/cvpaperchalleng}{Twitter @CVPaperChalleng (https://twitter.com/cvpaperchalleng)}, \href{http://www.slideshare.net/cvpaperchallenge}{SlideShare @cvpaper.challenge (http://www.slideshare.net/cvpaperchallenge)}}. Currently the project is run by around 20 members representing different organizations. We here describe our activities divided by (i) reading and (ii) writing papers.

\subsection{Reading papers}
Reading international conference papers clearly provides various advantages other than gaining an understanding of the current standing of your own research, such as acquiring ideas and methods used by researchers around the world.
%
%
We therefore undertook to extensively read papers, summarize them, and share them with others working in the same field. In the first year 2015, we focused on the IEEE Conference on Computer Vision and Pattern Recognition (CVPR) that is known as the top-tier conference in the field of computer vision, pattern recognition, and related fields. 
%
As the first step of this endeavor, we undertook to read all the 602 papers accepted during the CVPR2015~\cite{1_1,1_2,1_3,1_4,1_5,1_6,1_7,1_8,1_9,1_10,1_11,1_12,1_13,1_14,1_15,1_16,1_17,1_18,1_19,1_20,1_21,1_22,1_23,1_24,1_25,1_26,1_27,1_28,1_29,1_30,1_31,1_32,1_33,1_34,1_35,1_36,1_37,1_38,1_39,1_40,1_41,1_42,1_43,1_44,1_45,1_46,1_47,1_48,1_49,1_50,1_51,1_52,1_53,1_54,1_55,1_56,1_57,1_58,1_59,1_60,1_61,1_62,1_63,1_64,1_65,1_66,1_67,1_68,1_69,1_70,1_71,1_72,1_73,1_74,1_75,1_76,1_77,1_78,1_79,1_80,1_81,1_82,1_83,1_84,1_85,1_86,1_87,1_88,1_89,1_90,1_91,1_92,1_93,1_94,1_95,1_96,1_97,1_98,1_99,1_100,1_101,1_102,1_103,1_104,1_105,1_106,1_107,1_108,1_109,1_110,1_111,1_112,1_113,1_114,1_115,1_116,1_117,1_118,1_119,1_120,1_121,2_1,2_2,2_3,2_4,2_5,2_6,2_7,2_8,2_9,2_10,2_11,2_12,2_13,2_14,2_15,2_16,2_17,2_18,2_19,2_20,2_21,2_22,2_23,2_24,2_25,2_26,2_27,2_28,2_29,2_30,2_31,2_32,2_33,2_34,2_35,2_36,2_37,2_38,2_39,2_40,2_41,2_42,2_43,2_44,2_45,2_46,2_47,2_48,2_49,2_50,2_51,2_52,2_53,2_54,2_55,2_56,2_57,2_58,2_59,2_60,2_61,2_62,2_63,2_64,2_65,2_66,2_67,2_68,2_69,2_70,2_71,2_72,2_73,2_74,2_75,2_76,2_77,2_78,2_79,2_80,2_81,2_82,2_83,2_84,2_85,2_86,2_87,2_88,2_89,2_90,2_91,2_92,2_93,2_94,2_95,2_96,2_97,2_98,2_99,2_100,2_101,2_102,2_103,2_104,2_105,2_106,2_107,2_108,2_109,2_110,2_111,2_112,2_113,2_114,2_115,2_116,2_117,2_118,2_119,2_120,3_1,3_2,3_3,3_4,3_5,3_6,3_7,3_8,3_9,3_10,3_11,3_12,3_13,3_14,3_15,3_16,3_17,3_18,3_19,3_20,3_21,3_22,3_23,3_24,3_25,3_26,3_27,3_28,3_29,3_30,3_31,3_32,3_33,3_34,3_35,3_36,3_37,3_38,3_39,3_40,3_41,3_42,3_43,3_44,3_45,3_46,3_47,3_48,3_49,3_50,3_51,3_52,3_53,3_54,3_55,3_56,3_57,3_58,3_59,3_60,3_61,3_62,3_63,3_64,3_65,3_66,3_67,3_68,3_69,3_70,3_71,3_72,3_73,3_74,3_75,3_76,3_77,3_78,3_79,3_80,3_81,3_82,3_83,3_84,3_85,3_86,3_87,3_88,3_89,3_90,3_91,3_92,3_93,3_94,3_95,3_96,3_97,3_98,3_99,3_100,3_101,3_102,3_103,3_104,3_105,3_106,3_107,3_108,3_109,3_110,3_111,3_112,3_113,3_114,3_115,3_116,3_117,3_118,3_119,3_120,4_1,4_2,4_3,4_4,4_5,4_6,4_7,4_8,4_9,4_10,4_11,4_12,4_13,4_14,4_15,4_16,4_17,4_18,4_19,4_20,4_21,4_22,4_23,4_24,4_25,4_26,4_27,4_28,4_29,4_30,4_31,4_32,4_33,4_34,4_35,4_36,4_37,4_38,4_39,4_40,4_41,4_42,4_43,4_44,4_45,4_46,4_47,4_48,4_49,4_50,4_51,4_52,4_53,4_54,4_55,4_56,4_57,4_58,4_59,4_60,4_61,4_62,4_63,4_64,4_65,4_66,4_67,4_68,4_69,4_70,4_71,4_72,4_73,4_74,4_75,4_76,4_77,4_78,4_79,4_80,4_81,4_82,4_83,4_84,4_85,4_86,4_87,4_88,4_89,4_90,4_91,4_92,4_93,4_94,4_95,4_96,4_97,4_98,4_99,4_100,4_101,4_102,4_103,4_104,4_105,4_106,4_107,4_108,4_109,4_110,4_111,4_112,4_113,4_114,4_115,4_116,4_117,4_118,4_119,4_120,5_1,5_2,5_3,5_4,5_5,5_6,5_7,5_8,5_9,5_10,5_11,5_12,5_13,5_14,5_15,5_16,5_17,5_18,5_19,5_20,5_21,5_22,5_23,5_24,5_25,5_26,5_27,5_28,5_29,5_30,5_31,5_32,5_33,5_34,5_35,5_36,5_37,5_38,5_39,5_40,5_41,5_42,5_43,5_44,5_45,5_46,5_47,5_48,5_49,5_50,5_51,5_52,5_53,5_54,5_55,5_56,5_57,5_58,5_59,5_60,5_61,5_62,5_63,5_64,5_65,5_66,5_67,5_68,5_69,5_70,5_71,5_72,5_73,5_74,5_75,5_76,5_77,5_78,5_79,5_80,5_81,5_82,5_83,5_84,5_85,5_86,5_87,5_88,5_89,5_90,5_91,5_92,5_93,5_94,5_95,5_96,5_97,5_98,5_99,5_100,5_101,5_102,5_103,5_104,5_105,5_106,5_107,5_108,5_109,5_110,5_111,5_112,5_113,5_114,5_115,5_116,5_117,5_118,5_119,5_120,5_121}. We inherites the flow, however, in the second year 2016, we extend the covered range of conferences/journals (e.g. ICCV/ECCV/NIPS/PAMI/IJCV) in addition to the CVPR~\cite{1512_1,1512_2,1512_3,1512_4,1512_5,1512_6,1512_7,1512_8,1512_9,1512_10,1512_11,1512_12,1512_13,1512_14,1512_15,1512_16,1512_17,1512_18,1512_19,1512_20,1512_21,1512_22,1512_23,1512_24,1512_25,1512_26,1512_27,1512_28,1512_29,1512_30,1512_31,1512_32,1512_33,1512_34,1512_35,1512_36,1512_37,1512_38,1512_39,1512_40,1512_41,1512_42,1512_43,1512_44,1512_45,1512_46,1512_47,1512_48,1512_49,1512_50,1512_51,1512_52,1512_53,1512_54,1512_55,1512_56,1512_57,1512_58,1512_59,1512_60,1512_61,1512_62,1512_63,1512_64,1512_65,1512_66,1512_67,1512_68,1512_69,1512_70,1512_71,1512_72,1512_73,1512_74,1512_75,1512_76,1512_77,1512_78,1512_79,1512_80,1512_81,1512_82,1512_83,1512_84,1512_85,1512_86,1512_87,1512_88,1512_89,1512_90,1512_91,1512_92,1512_93,1512_94,1512_95,1512_96,1512_97,1512_98,1512_99,1512_100,1512_101,1512_102,1512_103,1512_104,1512_105,1512_106,1512_107,1512_108,1601_1,1601_2,1601_3,1601_4,1601_5,1601_6,1601_7,1601_8,1601_9,1601_10,1601_11,1601_12,1601_13,1601_14,1601_15,1601_16,1601_17,1601_18,1601_19,1601_20,1601_21,1601_22,1601_23,1601_24,1601_25,1601_26,1601_27,1602_1,1602_2,1602_3,1602_4,1602_5,1602_6,1602_7,1602_8,1602_9,1602_10,1602_11,1602_12,1602_13,1602_14,1602_15,1602_16,1602_17,1602_18,1602_19,1602_20,1602_21,1602_22,1602_23,1602_24,1602_25,1602_26,1602_27,1602_28,1602_29,1602_30,1602_31,1602_32,1602_33,1602_34,1602_35,1602_36,1602_37,1602_38,1602_39,1602_40,1602_41,1602_42,1602_43,1602_44,1602_45,1602_46,1602_47,1602_48,1602_49,1602_50,1602_51,1603_1,1603_2,1603_3,1603_4,1603_5,1603_6,1603_7,1603_8,1603_9,1603_10,1603_11,1603_12,1603_13,1603_14,1603_15,1603_16,1603_17,1603_18,1603_19,1603_20,1603_21,1603_22,1603_23,1603_24,1603_25,1603_26,1603_27,1603_28,1603_29,1603_30,1603_31,1603_32,1603_33,1603_34,1603_35,1603_36,1603_37,1603_38,1603_39,1603_40,1603_41,1603_42,1603_43,1603_44,1603_45,1603_46,1603_47,1603_48,1603_49,1603_50,1603_51,1603_52,1603_53,1603_54,1603_55,1603_56,1603_57,1603_58,1603_59,1603_60,1603_61,1603_62,1603_63,1603_64,1603_65,1603_66,1603_67,1603_68,1603_69,1603_70,1604_1,1604_2,1604_3,1604_4,1604_5,1604_6,1604_7,1604_8,1604_9,1604_10,1604_11,1604_12,1604_13,1604_14,1604_15,1604_16,1604_17,1604_18,1604_19,1604_20,1604_21,1604_22,1604_23,1604_24,1604_25,1604_26,1604_27,1604_28,1604_29,1604_30,1604_31,1604_32,1604_33,1604_34,1604_35,1604_36,1604_37,1604_38,1604_39,1604_40,1604_41,1604_42,1604_43,1604_44,1604_45,1604_46,1604_47,1604_48,1604_49,1604_50,1604_51,1604_52,1604_53,1604_54,1604_55,1604_56,1604_57,1604_58,1604_59,1604_60,1604_61,1604_62,1604_63,1604_64,1604_65,1604_66,1605_1,1605_2,1605_3,1605_4,1605_5,1605_6,1605_7,1605_8,1605_9,1605_10,1605_11,1605_12,1605_13,1605_14,1605_15,1605_16,1605_17,1605_18,1605_19,1605_20,1605_21,1605_22,1605_23,1605_24,1605_25,1605_26,1605_27,1605_28,1605_29,1605_30,1605_31,1605_32,1605_33,1605_34,1605_35,1605_36,1605_37,1605_38,1605_39,1605_40,1605_41,1605_42,1605_43,1605_44,1605_45,1605_46,1605_47,1605_48,1605_49,1605_50,1605_51,1605_52,1605_53,1605_54,1605_55,1605_56,1605_57,1605_58,1605_59,1605_60,1605_61,1605_62,1605_63,1605_64,1605_65,1605_66,1605_67,1605_68,1605_69,1605_70,1605_71,1605_72,1605_73,1605_74,1605_75,1605_76,1605_77,1605_78,1605_79,1605_80,1605_81,1605_82,1605_83,1605_84,1605_85,1605_86,1605_87,1605_88,1605_89,1605_90,1605_91,1605_92,1605_93,1606_1,1606_2,1606_3,1606_4,1606_5,1606_6,1606_7,1606_8,1606_9,1606_10,1606_11,1606_12,1606_13,1606_14,1606_15,1606_16,1606_17,1606_18,1606_19,1606_20,1606_21,1606_22,1606_23,1606_24,1606_25,1606_26,1606_27,1606_28,1606_29,1606_30,1606_31,1606_32,1606_33,1606_34,1606_35,1606_36,1606_37,1606_38,1606_39,1606_40,1606_41,1606_42,1606_43,1606_44,1606_45,1606_46,1606_47,1606_48,1606_49,1606_50,1606_51,1606_52,1606_53,1606_54,1606_55,1606_56,1606_57,1606_58,1606_59,1606_60,1606_61,1606_62,1606_63,1606_64,1606_65,1606_66,1606_67,1606_68,1606_69,1606_70,1606_71,1606_72,1606_73,1606_74,1606_75,1606_76,1606_77,1606_78,1606_79,1606_80,1606_81,1606_82,1606_83,1606_84,1606_85,1606_86,1606_87,1606_88,1606_89,1606_90,1606_91,1606_92,1606_93,1606_94,1606_95,1606_96,1606_97,1606_98,1606_99,1606_100,1606_101,1606_102,1606_103,1606_104,1606_105,1606_106,1606_107,1606_108,1606_109,1606_110,1606_111,1606_112,1606_113,1606_114,1606_115,1606_116,1606_117,1606_118,1606_119,1606_120,1606_121,1606_122,1606_123,1606_124,1606_125,1606_126,1606_127,1606_128,1606_129,1606_130,1606_131,1606_132,1606_133,1606_134,1606_135,1606_136,1606_137,1606_138,1606_139,1606_140,1606_141,1606_142,1606_143,1606_144,1606_145,1606_146,1606_147,1606_148,1606_149,1606_150,1606_151,1606_152,1606_153,1607_1,1607_2,1607_3,1607_4,1607_5,1607_6,1607_7,1607_8,1607_9,1607_10,1607_11,1607_12,1607_13,1607_14,1607_15,1607_16,1607_17,1607_18,1607_19,1607_20,1607_21,1607_22,1607_23,1607_24,1607_25,1607_26,1607_27,1607_28,1607_29,1607_30,1607_31,1607_32,1607_33,1607_34,1607_35,1607_36,1607_37,1607_38,1607_39,1607_40,1607_41,1607_42,1607_43,1607_44,1607_45,1607_46,1607_47,1607_48,1607_49,1607_50,1608_1,1608_2,1608_3,1608_4,1608_5,1608_6,1608_7,1608_8,1608_9,1608_10,1608_11,1608_12,1608_13,1608_14,1608_15,1608_16,1608_17,1608_18,1608_19,1608_20,1608_21,1608_22,1608_23,1608_24,1609_1,1609_2,1609_3,1609_4,1609_5,1609_6,1609_7,1609_8,1609_9,1609_10,1609_11,1609_12,1609_13,1609_14,1609_15,1609_16,1609_17,1609_18,1609_19,1609_20,1609_21,1609_22,1609_23,1609_24,1609_25,1609_26,1609_27,1609_28,1609_29,1609_30,1609_31,1609_32,1609_33,1609_34,1609_35,1609_36,1609_37,1609_38,1609_39,1609_40,1609_41,1610_1,1610_2,1610_3,1610_4,1610_5,1610_6,1610_7,1610_8,1610_9,1610_10,1610_11,1610_12,1610_13,1610_14,1610_15,1610_16,1610_17,1610_18,1610_19,1610_20,1610_21,1610_22,1610_23,1610_24,1610_25,1610_26,1610_27,1610_28,1610_29,1611_1,1611_2,1611_3,1611_4,1611_5,1611_6,1611_7,1611_8,1611_9,1611_10,1611_11,1611_12,1611_13,1611_14,1611_15,1611_16,1611_17,1611_18,1611_19,1611_20,1611_21,1611_22,1611_23,1611_24,1611_25,1611_26,1611_27,1611_28,1611_29,1611_30,1611_31,1611_32,1611_33,1611_34,1611_35,1611_36,1611_37,1611_38,1611_39,1611_40,1611_41,1611_42,1611_43,1611_44,1611_45,1611_46,1611_47,1611_48,1611_49,1611_50,1611_51,1611_52,1611_53,1611_54,1611_55,1611_56,1611_57,1612_1,1612_2,1612_3,1612_4,1612_5,1612_6,1612_7,1612_8,1612_9,1612_10,1612_11,1612_12,1612_13,1612_14,1612_15,1612_16,1612_17,1612_18,1612_19,1612_20,1612_21,1612_22,1612_23,1612_24,1612_25,1612_26,1612_27,1612_28,1612_29,1612_30,1612_31,1612_32,1612_33,1612_34,1612_35,1612_36,1612_37,1612_38,1612_39,1612_40,1612_41,1612_42,1612_43,1612_44,1612_45,1612_46,1612_47,1612_48,1612_49,1612_50,1612_51,1612_52,1612_53,1612_54,1612_55,1612_56,1612_57,1612_58,1612_59,1612_60,1612_61,1612_62,1612_63,1612_64,1612_65,1612_66,1612_67,1612_68,1612_69,1612_70,1612_71,1612_72,1612_73,1612_74,1612_75,1612_76,1612_77,1612_78,1612_79,1612_80,1612_81,1612_82,1612_83,1612_84,1612_85,1612_86,1612_87,1612_88,1612_89,1612_90,1612_91,1612_92,1612_93,1612_94,1612_95,1612_96,1612_97,1612_98,1612_99,1612_100,1612_101,1612_102,1612_103,1612_104,1612_105,1612_106,1612_107,1612_108,1612_109,1612_110,1612_111,1612_112,1612_113,1612_114,1612_115,1612_116,1612_117,1612_118,1612_119,1612_120,1612_121,1612_122,1612_123,1612_124,1612_125,1612_126,1612_127,1612_128,1612_129,1612_130,1612_131,1612_132,1612_133,1612_134,1612_135,1612_136,1612_137,1612_138,1612_139,1612_140,1612_141,1612_142,1612_143,1612_144,1612_145,1612_146,1612_147,1612_148,1612_149,1612_150,1612_151,1612_152,1612_153,1612_154,1612_155,1612_156,1612_157,1612_158,1612_159,1612_160,1612_161,1612_162,1612_163,1612_164,1612_165,1612_166,1612_167,1612_168,1612_169,1612_170,1612_171,1612_172,1612_173,1612_174,1612_175,1612_176,1612_177,1612_178,1612_179,1612_180,1612_181,1612_182,1612_183,1612_184,1612_185,1612_186,1612_187,1612_188,1612_189,1612_190,1612_191,1612_192,1612_193,1612_194,1612_195,1612_196,1612_197,1612_198,1612_199,1612_200,1612_201,1612_202,1612_203,1612_204,1612_205,1612_206,1612_207,1612_208,1612_209,1612_210,1612_211,1612_212,1612_213,1612_214,1612_215,1612_216,1612_217,1612_218,1612_219,1612_220,1612_221,1612_222,1612_223,1612_224,1612_225,1612_226,1612_227,1612_228,1612_229,1612_230,1612_231,1612_232,1612_233,1612_234,1612_235,1612_236,1612_237,1612_238,1612_239,1612_240,1612_241,1612_242,1612_243,1612_244}. We have comprehensively reviewed in the computer vision and related works to propose sophisticated ideas into the research fields. 

\subsection{Writing papers}

We have proposed a mechanism for the systematization of knowledge, the generation of sophisticated ideas, and as well as the writing papers especially in new research problems~\cite{cvpaper.challengearXiv2016}. The mechanism is composed of the following:

\begin{enumerate}
    \item Acquiring knowledge: In the case we have read 1,600+ papers
    \item From knowledge to ideas: Individually generate ideas
    \item Consolidation of ideas: Group discussion
    \item Sharpen the ideas: Iteratively conduct 2 and 3
    \item From ideas to a sophisticated research theme: Doctors distillates research themes
    \item Consideration: Rethinking themes by implementation
    \item Research: Do experiments
    \item Output: Publish papers
\end{enumerate}

\section{Futuristic computer vision}

The paper predicts futuristic computer vision and introduces our researches above the line. We list the 9 topics: 

\begin{enumerate}
\item Pure motion representation in video
\item Generalizing 3D representation
\item World-level recognition and reconstruction
\item Fully-automatic dataset creation with WWW, CG, automation
\item Transforming and understanding physical quantity
\item Looking at latent knowledge in an image
\item Self-supervised learning
\item (AI) History repeats itself
\item The pixel redefinition 
\end{enumerate}

\subsection{Pure motion representation in video}

In the past few years, convolutional neural networks have made great contributions to image recognition. Undoubtedly, the current area of interest has been shifting a temporal analysis of video input, such as action recognition and motion representation. These techniques are expected to be useful in such areas as visual surveillance, autonomous driving, and VR/AR. Various video recognition frameworks have been proposed, but as yet, there is no (enough) de facto standard for the field of motion representation. To tackle this urgent issue, the research field has tried various approaches, including conducting a workshop to elicit ideas for representing motion ~\cite{FernandoECCVW2016}.

In this era of deep learning, the two-stream CNN~\cite{SimonyanNIPS2014} has replaced most of the hand-crafted approaches used on human action datasets. Two-stream CNN efficiently categorizes human actions using the RGB image (spatial stream) and flow image (temporal stream), and it performs better than either DT or IDT.  We note that use of the spatial-stream achieved over 70\% on the UCF101 dataset~\cite{UCF101}; that is, the per-frame RGB feature was able to categorize the 101 action classes without a temporal feature. The action datasets mainly occupies background areas against to the human areas in the image sequences. This still shows that there is a problem with background dependency, and it is important to find a more sophisticated way to represent motion. Here we introduce our ``Human Action Recognition without Human"~\cite{HeECCVW2016} and ``Motion Representation with Acceleration Images"~\cite{KataokaECCVW2016}. Especially in the ``Human Action Recognition without Human", we have validated the background effects in video recognition. Moreover in the ``Motion Representation with Acceleration Images", we propose an additional motion representation from a conventional descriptor.

\begin{figure}[t] 
\begin{center}
   \includegraphics[width=1.0\linewidth]{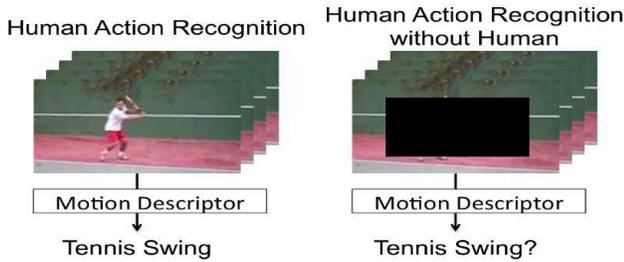}
\end{center}
   \caption{Human action recognition (left) and human action
recognition without human (right): We simply replace
the center-around area with a black background in an image
sequence. We evaluate the performance rate with only the
limited background sequence as a contextual cue.}
\label{fig:withouthuman}
\end{figure}

\begin{figure}[t] 
\begin{center}
   \includegraphics[width=1.0\linewidth]{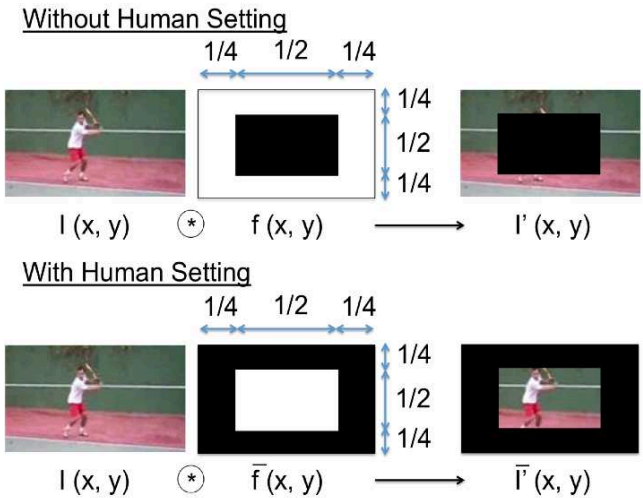}
\end{center}
   \caption{Image filtering for human action recognition without
human.}
    \label{fig:filter}
\end{figure}

{\bf Human Action Recognition without Human~\cite{HeECCVW2016}} Motion representation is frequently discussed in human action recognition. We have examined several sophisticated options, such as dense trajectories (DT) and the two-stream convolutional neural network (Two-Stream CNN). However, some features from the background could be too strong, as shown in some recent studies on human action recognition (for example, the spatial-stream in Two-Stream CNN can recognize 70 \% on UCF101 without temporal information). Therefore, we considered whether a background sequence alone can classify human actions in current large-scale action datasets (e.g., UCF101). In the research, we propose a novel concept for human action analysis that is named ``human action recognition without human" (see Figure~\ref{fig:withouthuman}). An experiment clearly shows the effect of a background sequence for understanding an action label. The setting is shown in Figure~\ref{fig:filter}. 

To the best of our knowledge, this is the first study of human action recognition without human. However, we should not have done that kind of thing. The motion representation from a background sequence is effective to classify videos in a human action database. We demonstrated human action recognition in with and without a human settings on the UCF101 dataset. The results show the setting without a human (47.42 \%; without human setting) was close to the setting with a human (56.91\%; with human setting). We must accept this reality to realize better motion representation.

\begin{figure}[t] 
\begin{center}
   \includegraphics[width=1.0\linewidth]{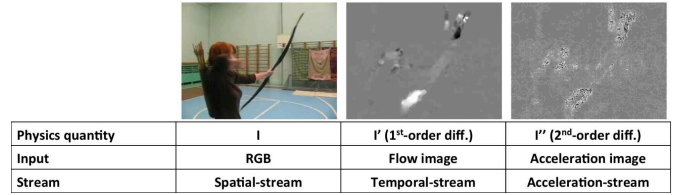}
\end{center}
   \caption{Image representation of RGB (I), flow (I'), and acceleration (I'').}
    \label{fig:acceleration_input}
\end{figure}

{\bf Motion Representation with Acceleration Images~\cite{KataokaECCVW2016}} Referring to the~\cite{HeECCVW2016}, we propose the simple but effective representation of using ``acceleration images'' to represent a change of a flow image. The Figure~\ref{fig:acceleration_input} shows the visual comparison. The acceleration images must be significant because the representation is different from position (RGB) and speed (flow) images. We apply two-stream CNN as the baseline; then, we employ an acceleration stream in addition to the spatial and temporal streams. The acceleration images are generated by differential calculations from a sequence of flow images. Although the sparse representation tends to be noisy data (see Figure~\ref{fig:acceleration_input}), automatic feature learning with CNN can significantly pick up a necessary feature in the acceleration images.

\subsection{Generalizing 3D representation}

We believe that a 3-dimensional $XYZ$ data must be further improved for real-world understanding. The conventional approaches such as SpinImages~\cite{JohnsonICRA1997}, SHOT~\cite{TombariECCV2010} have accelerated recognition performance in 3D point clouds. Moreover the Multi-View CNN (MVCNN)~\cite{SuICCV2015} and Deep Sliding Shapes (DSS)~\cite{SongCVPR2016} have been employed in the age of deep learning. The MVCNN proposed a view-free recognition through per-view learning and view-pooling from 12 viewpoints. The DSS executed both 3D object proposal and 2D-3D object classification in 3D space.

However, the current 3D object recognition is focusing on so-called specific object recognition not about the general 3D object recognition. The breakthrough in 3D object recognition may occur if a model learns flexible models which generalize 3D object shapes like ImageNet pre-trained model in 2D image recognition~\cite{DonahueICML2014}. The pre-trained model is also strong at transfer learning with activation features. Here we must try to propose a bigger database and sophisticated model in addition to the recent frameworks. By comparing the ShapeNet Core-55 (3D-DB) with the ImageNet (2D-DB), we must collect a bigger scale 3D data to train a sophisticated model for general 3D object recognition. The architecture should be improved.

\subsection{World-level recognition and reconstruction}

\begin{figure*}[t]
\centering
\subfigure[Fashion trend changes in New York, Paris, London, and Tokyo, 2014--2016. Our goal is to reveal the latest trends for each year.]{\includegraphics[width=0.35\linewidth]{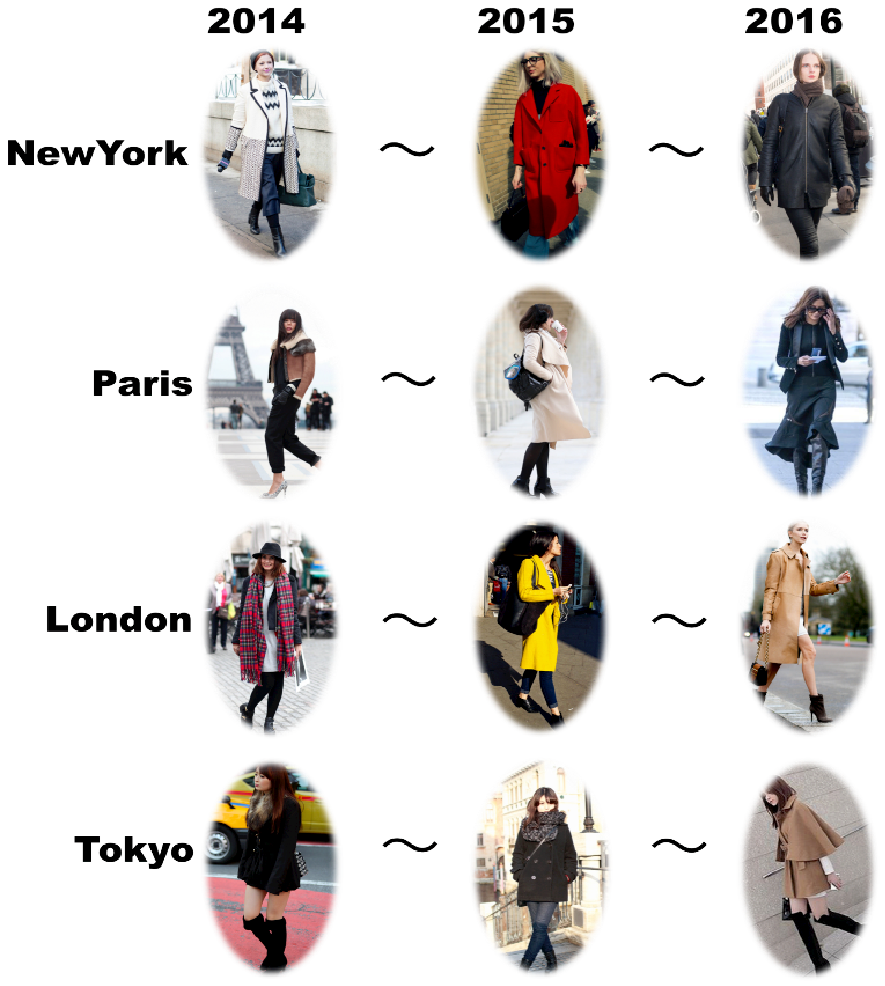}
\label{fig:trends}}
\subfigure[Fashion culture database (FCDB; ours)]{\includegraphics[width=0.63\linewidth]{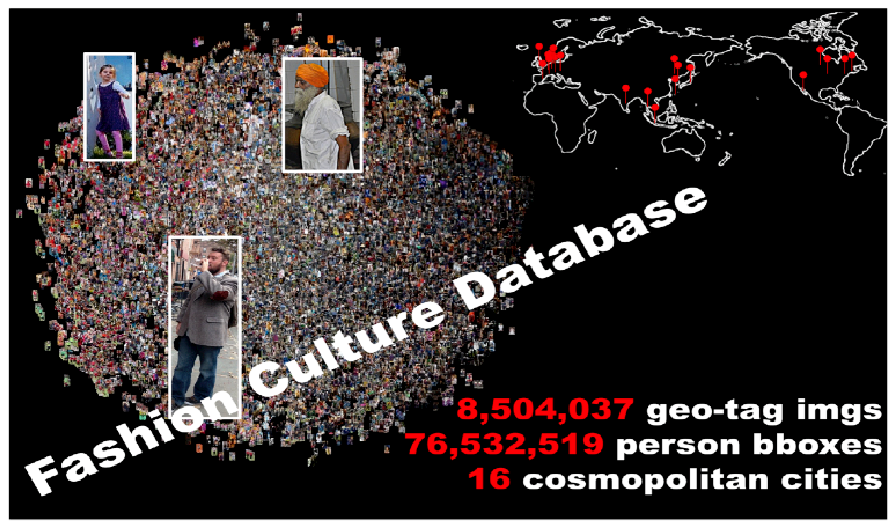}
\label{fig:fcdb_abst}}
\caption{Worldwide fashion change analysis from the self-collected fashion culture database (FCDB). We analyze changing fashion cultures for every season (a). The images are taken from Flickr, under a Creative Commons license. The FCDB consists of 8M geo-tagged images and 76M person bounding boxes in 16 cosmopolitan cities (b).}
\label{fig:concept}
\end{figure*}

Recently, the world-level analysis is being done with geo-tagged images from web-photo and social networking servises. The geo-tagged images include GPS information, time-stamp in addition to the image itself. We have confirmed a couple of innovative algorithms by combining geo-tagged information and image data. One of the achievement is undoubtedly ``Building Rome in a Day~\cite{AgarwalICCV2009}" with structure-from-motion (SfM). The recent SfM problem is implemented wider area in ``Reconstructing the World in Six Days~\cite{HeinlyCVPR2015}". The kinds of 3D reconstruction have developed vision-based algorithm such as keypoint matching and bundle adjustment. 

On one hand, an integrated method both scene recognition and geo-tagged image analysis is proposed in City Perception~\cite{ZhouECCV2014}. The city-scale analysis is based on the scene recognition with stacked likelihood from the PlaceCNN~\cite{ZhouNIPS2014}. 

We can improve the world-level reconstruction and recognition since the world-level 3D reconstruction stays in certain landmarks estimation, and the city perception focuses arbitrary images at each city. One of the causes is the dataset scale, in which the city perception has only a couple of million to treat world-level recognition. One more reason is the lack of semantic labeling (e.g. object detection, semantic segmentation) in an image. The recent semantic labeling effectively helps to enhance concept-level annotations such as object categories and building attributes. In the scene parsing, the ADE20K dataset~\cite{ZhouarXiv2016} is proposed in the ILSVRC2016 to build up detailed scene- and object-level segmentations which contain 150 categories. Moreover, we can get $10^7$-order images from Web, SNS and map data. What can we estimate from $10^7$ geo-tagged images and detailed scene parsing?

We here introduce our ``Changing Fashion Cultures"~\cite{AbearXiv2017} to analyze world-wide fashion trends (see Figure~\ref{fig:concept}). 

\begin{figure}[t] 
\begin{center}
   \includegraphics[width=1.0\linewidth]{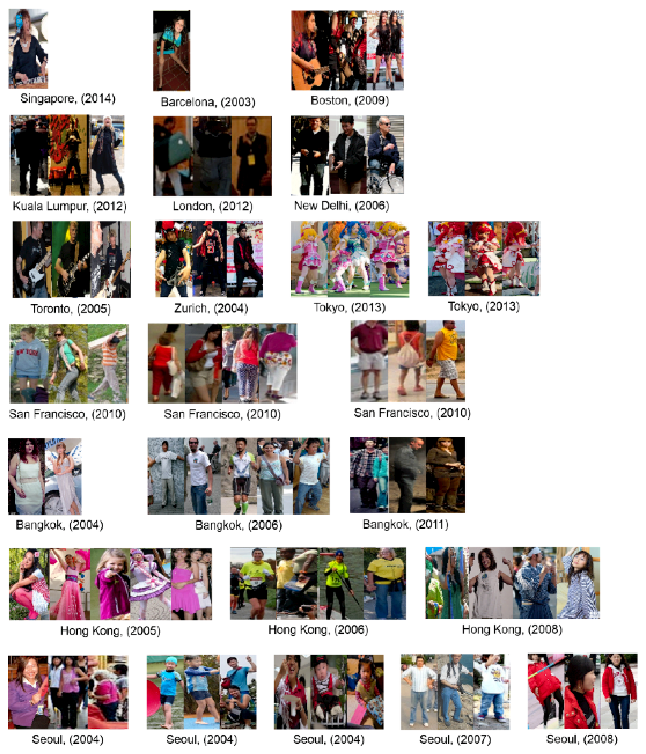}
\end{center}
   \caption{Visualization of fashion trends: The figure shows randomly selected cutting-edge fashion trends per city and year. For example, Tokyo (2013) shows one of the cutting-edge fashion trend at Tokyo in 2013. Although the FTD is not perfect, we confirm the concept of “changing fashion culture” is achieved as the result of FCDB collection and unsupervised analyzer.}
    \label{fig:trends}
\end{figure}

\begin{figure}[t] 
\begin{center}
   \includegraphics[width=1.0\linewidth]{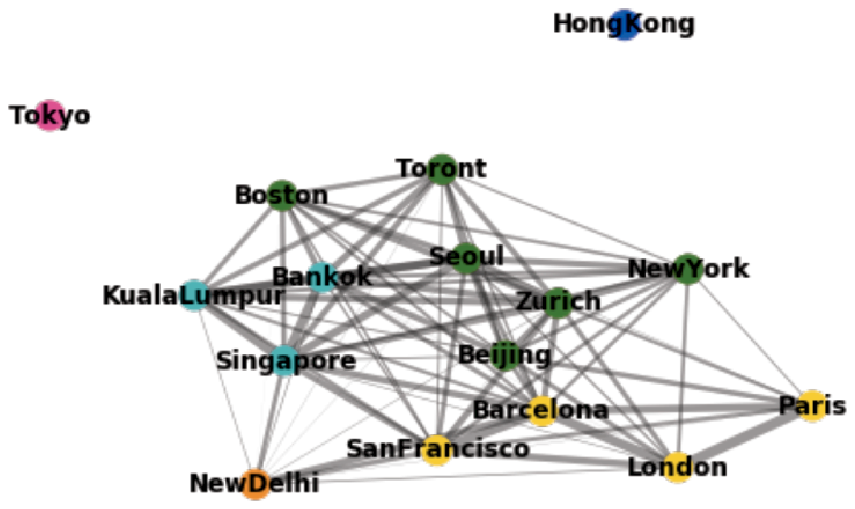}
\end{center}
   \caption{Fashion-based city similarity graph: The input is from 1,000-dim codeword vectors with StyleNet+Lab. The nodes and edges correspond to cities and similarities between pairs of cities, where the line thickness indicates the degree of similarity. In the graph representation, we set the thresholding value as 0.2 in order to eliminate low correlations.}
    \label{fig:citysim}
\end{figure}

{\bf Changing Fashion Cultures~\cite{AbearXiv2017}} The research presents a novel concept that analyzes and visualizes worldwide fashion trends. Our goal is to reveal cutting-edge fashion trends without displaying an ordinary fashion style. To achieve the fashion-based analysis, we created a new fashion culture database (FCDB), which consists of 76 million geo-tagged images in 16 cosmopolitan cities. By grasping a fashion trend of mixed fashion styles, the paper also proposes an unsupervised fashion trend descriptor (FTD) using a fashion descriptor, a codeword vetor, and temporal analysis. To unveil fashion trends in the FCDB, the temporal analysis in FTD effectively emphasizes consecutive features between two different times. In experiments, we clearly show the analysis of fashion trends (see Figure~\ref{fig:trends}) and fashion-based city similarity (see Figure~\ref{fig:citysim}). As the result of large-scale data collection and an unsupervised analyzer, the proposed approach achieves world-level fashion visualization in a time series.

\subsection{Fully-automatic dataset creation with WWW, CG, automation}

What if we were able to automatically generate a large-scale database like ImageNet? There are some fully-automated database such as YFCC100M~\cite{ThomeeACM2016} (flickr style), CG pedestrian data~\cite{HattoriCVPR2015} (computer graphics) and iLab-20M~\cite{BorjiCVPR2016} (automation technology) regardless of w/ or w/o object-level categories. Hereafter the increase of fully-automated dataset is expected. More recent research, for example, the generative adversarial nets (GANs)~\cite{GoodfellowNIPS2014} achieved image generation close to real data. Style transfer (e.g. photo transfer~\cite{LuanarXiv2017}) has also been developing realistic data generation from image pairs. 

Now we are cooperating annotation tasks with computers to create an image database~\cite{RussakovskyCVPR2015,PapadopoulosCVPR2016}. To alleviate human burden, a computer (i) ask a question to human~\cite{RussakovskyCVPR2015}, (ii) proposal detection of ground truth~\cite{PapadopoulosCVPR2016}. Hereafter the framework improves dataset quality, and we expect that the dataset creation is done by fully-automatic annotation. There are a couple of annotation techniques such as robot hands, automation, simulator and gaming for dataset creation. Although the current work is being developing, an image~\cite{GoodfellowNIPS2014}/video~\cite{VondrickNIPS2016} generator (e.g. generative adversarial nets) create a realistic data. Near future we will get a wider and bigger XYZ data with the kind of generator.

\subsection{Transforming and understanding physical quantity}

Thanks to the recent ConvNet and RecurrentNet, a physical transformation from image sequence to audio is achieved in~\cite{OwensCVPR2016}. The physical quantity transformation called ``Visually Indicated Sounds" mapped video to audio. Although the technique combined a simple CNN+LSTM combined model, a generated sound cannot be easily classified (cannot be done even by a human). This is the first example that passed a sound turing test. 



\subsection{Looking at latent knowledge in an image}

In the topic we have published a couple of papers such as ``Academy Award Prediction"~\cite{MatsuzakiMVA2017}, ``Transitional Action Recognition"~\cite{KataokaBMVC2016}.

{\bf Academy Award Prediction~\cite{MatsuzakiMVA2017}} The research aims at estimating the winner of world-wide film festival from the exhibited movie poster. The task is an extremely challenging because the estimation must be done with only an exhibited movie poster, without any film ratings and box-office takings. In order to tackle this problem, we have created a new database which is consist of all movie posters included in the four biggest film festivals. The movie poster database (MPDB) contains historic movies over 80 years which are nominated a movie award at each year. We apply a couple of feature types, namely hand-craft, mid-level and deep feature to extract various information from a movie poster. Our experiments showed suggestive knowledge, for example, the Academy award estimation can be better rate with a color feature and a facial emotion feature generally performs good rate on the MPDB. The research may suggest a possibility of modeling human taste for a movie recommendation.

\begin{figure}[t]
\begin{center}
 \includegraphics[width=1.0\linewidth]{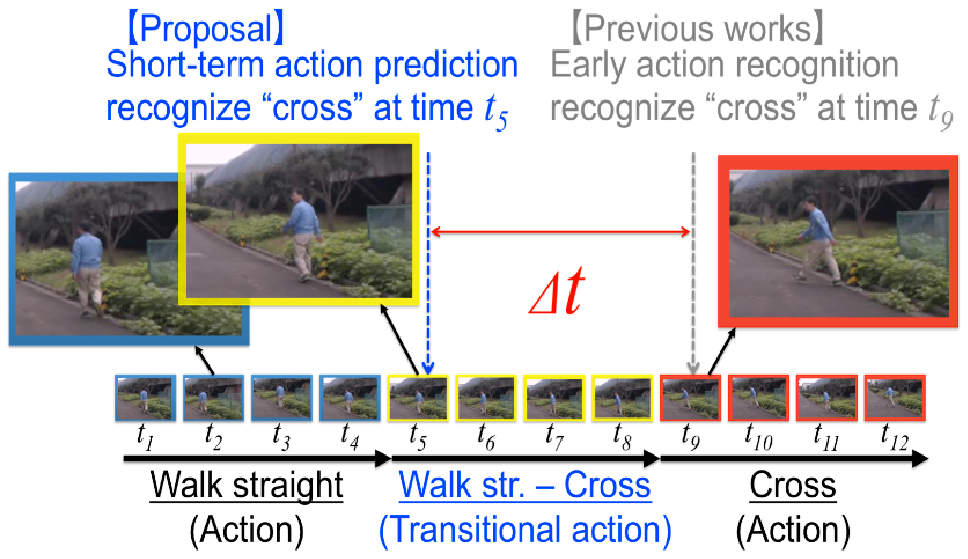}
\end{center}
\caption{Recognition of transitional actions for short-term action prediction: Our proposal adds a transitional action class \textit{Walk straight - cross} between \textit{Walk straight} and \textit{cross}. Identification of transitional actions allow us to understand the next activity at time \textit{$t_{5}$} before an early action recognition approach at time \textit{$t_{9}$}.}
\label{fig:problem_tar}
\end{figure}

\begin{figure*}[t]
\begin{center}
 \includegraphics[width=0.8\linewidth]{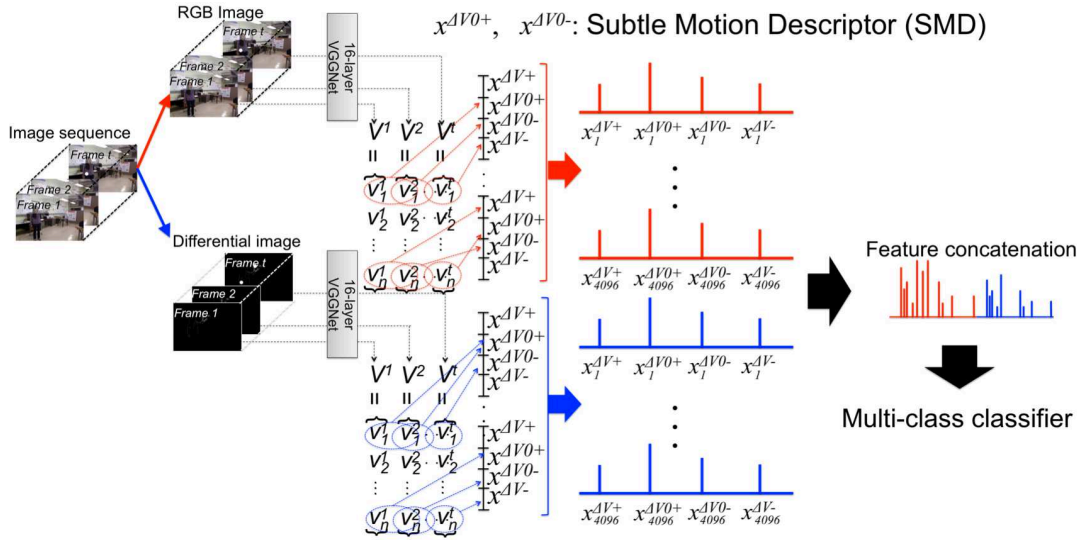}
\end{center}
 \caption{Discriminative temporal CNN feature with SMD for transitional action recognition: Multi-channel input from RGB and differential image is divided into two streams. At each frame, a CNN-based feature ($V^{t}$) is extracted with the first fully connected layer of 16-layer VGGNet ($N=4{,}096$). The consecutive subtractions ($\Delta V^{t}$) are pooled into four vectors, namely $x^{\Delta V^{+}}, x^{\Delta V^{0^{+}}}, x^{\Delta V^{-}}, x^{\Delta V^{0^{-}}}$. Here, the $x^{\Delta V^{0^{+}}}$ and $x^{\Delta V^{0^{-}}}$ are the proposed SMD. The feature concatenation of RGB and differential image streams is the final classification vector.}
\label{fig:framework_smd}
\end{figure*}

\begin{figure*}[t]
\begin{center}
   \includegraphics[width=0.75\linewidth]{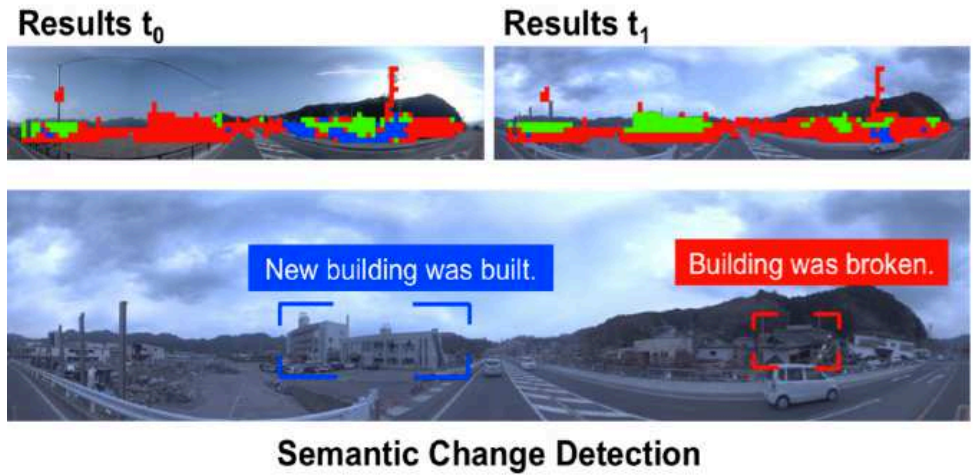}
\end{center}
   \caption{Concept of semantic change detection}
\label{fig:scd}
\end{figure*}

\begin{figure}[t]
\begin{center}
   \includegraphics[width=0.80\linewidth]{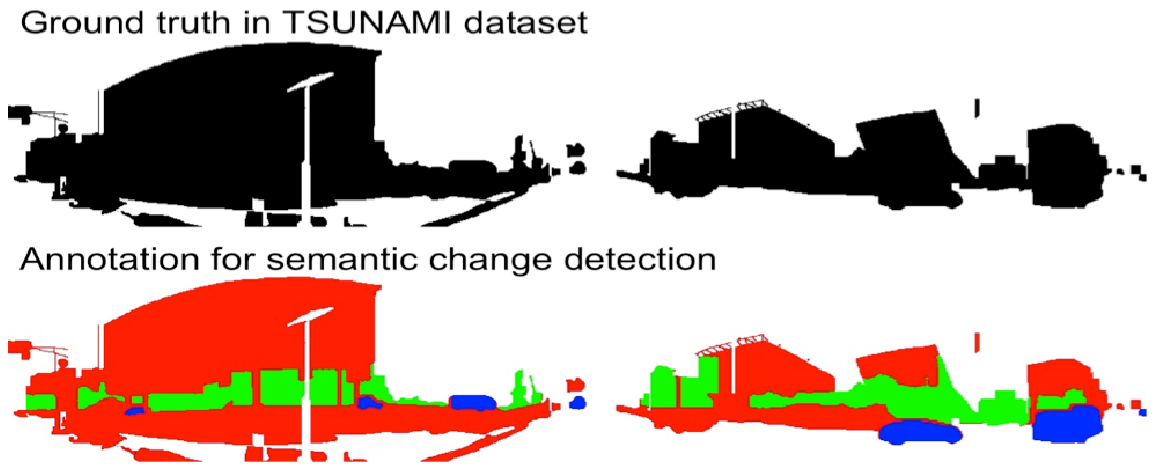}
\end{center}
   \caption{Original annotation in the TSUNAMI dataset~\cite{1512_96} (top) and semantic change detection annotation (bottom): Three labels were inserted into the dataset: car (blue), building (green), and rubble (red)}
\label{fig:annotation_scd}
\end{figure}

{\bf Semantic Change Detection~\cite{KataokaarXiv2016_scd}}
Change detection is the study of detecting changes between two different images of a scene taken at different times. By the detected change areas, however, a human cannot understand how different the two images. Therefore, a semantic understanding is required in the change detection research such as disaster investigation. The research proposes the concept of \textit {semantic change detection}, which involves intuitively inserting semantic meaning into detected change areas. The concept is shown in Figure~\ref{fig:scd}. We mainly focus on the novel semantic segmentation in addition to a conventional change detection approach. In order to solve this problem and obtain a high-level of performance, we propose an improvement to the hypercolumns representation~\cite{1_49}, hereafter known as hypermaps, which effectively uses convolutional maps obtained from CNN. We also employ multi-scale feature representation captured by different image patches. We applied our method to the TSUNAMI Panoramic Change Detection dataset~\cite{1512_96}, and re-annotated the changed areas of the dataset via semantic classes (see Figure~\ref{fig:annotation_scd}). The results show that our multi-scale hypermaps provided outstanding performance on the re-annotated TSUNAMI dataset.

{\bf Transitional Action Recognition~\cite{KataokaBMVC2016}}
We address \textbf{transitional actions} class as a class between actions (see Figure~\ref{fig:problem_tar}). Transitional actions should be useful for producing \textbf{short-term action predictions} while an action is transitive. However, transitional action recognition is difficult because actions and transitional actions partially overlap each other. To deal with this issue, we propose a \textbf{subtle motion descriptor (SMD)} that identifies the sensitive differences between actions and transitional actions (see Figure~\ref{fig:framework_smd}). The two primary contributions in this research are as follows: (i) defining transitional actions for short-term action predictions that permit earlier predictions than early action recognition, and (ii) utilizing convolutional neural network (CNN) based SMD to present a clear distinction between actions and transitional actions. Using three different datasets, we will show that our proposed approach produces better results than do other state-of-the-art models. The experimental results clearly show the recognition performance effectiveness of our proposed model,  as well as its ability to comprehend temporal motion in transitional actions.

\subsection{Self-supervised learning}

We should learn from the remarkable self-supervised learning like picking robots~\cite{LevinearXiv2016} and AlphaGo~\cite{SilverNature2016}. However, the learning systems operate only in a limited enviroment. In the next step we must extend the learning systems. For further explanation about 2.7, see other papers.

\subsection{(AI) History repeats itself}

\begin{figure}[t]
\begin{center}
   \includegraphics[width=0.9\linewidth]{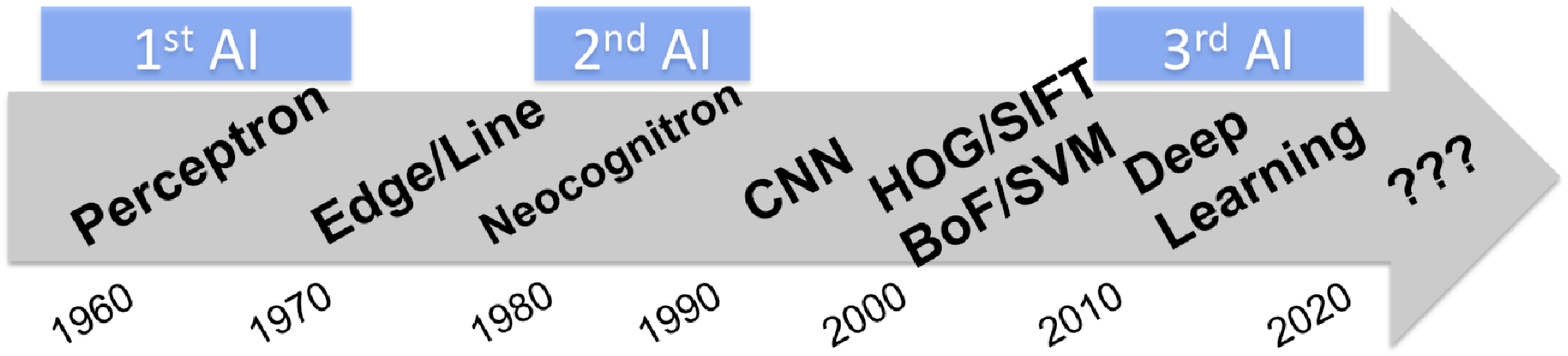}
\end{center}
   \caption{The $1^{st}$, $2^{nd}$ and $3^{rd}$ AI booms and winters: ($1^{st}$ boom) perceptron, ($1^{st}$ winter) edge/line detection, ($2^{nd}$ boom) neocognitron, ($2^{nd}$ winter) HOG/SIFT/BoF/SVM, ($3^{rd}$ boom [now]) deep learning. Here we anticipate a next framework after $3^{rd}$ AI.}
\label{fig:ai}
\end{figure}

The use of neural net and hand-crafted feature is repeated in the computer vision field (see Figure~\ref{fig:ai}). The suggestive knowledge motivates us to study an advanced hand-crafted feature after the DCNN in the recent $3^{rd}$ AI. We anticipate that the hand-crafted feature should collaborate with deeply learned parameters since an outstanding performance is achieved with the automatic feature learning.

The $1^{st}$ AI has started from the perceptron~\cite{Rosenblatt1961} which is the basic theory in neural networks. The (simple) perceptron is constructed by 2-layer with input and output layers. The iteration of AI booms and winters (e.g. edge detection~\cite{CannyMIT1983}, Neocognitron~\cite{FukushimaBC1980}, back prop.~\cite{RumelhartNature1986}, CNN~\cite{LeCunIEEE1998}, SIFT~\cite{LoweIJCV2004}, HOG~\cite{DalalCVPR2005}, Deep Learning~\cite{KrizhevskyNIPS2012}) may suggest us to propose the next generation algorithms.

At each AI boom, combined framework with genetic algorithm (GA) and neural net is coming to improve the parameter and architecture. The GA was proposed in 1970s~\cite{HollandMIT1975} and improved the framework in~\cite{Eshelman1991}. In the current AI boom, the GA framework is employed into deep neural networks~\cite{RealarXiv2017}. They applied several components and parameters such as convolution, pooling, batch normalization to adaptively learn an optimized architecture. The evolutional architecture achieved a comparative rate to the Wide ResNet~\cite{1606_112} on the CIFAR dataset.

What is the next movement? According to the previous stream, hand-crafted feature will be occurred. However, the current deep learned representation is enough strong, the DNN framework can exclude a rise of an improved hand-crafted feature. Therefore how about extending the space from 2D image to 3D volume data, like xyz (real space) or xyt (video stream)? We have only very early standard approaches such as PointNet~\cite{QiCVPR2017,QiarXiv2017} and C3D~\cite{1605_72}. Does the deep learning continue to be used as it is, or the local features are revisited? This flow is remarkable. Of course we must keep to propose the novel frameworks, not limited to these two.

\subsection{The pixel redefinition}

Camera obscura is said to be the origin of the camera~\cite{Haytham1021}. A photo is manually recorded by an image projected on a wall. The current RGB-image system computationally imitates the camera obscura. However, is this mechanism still useful in the era of deep learning? The digital image recognition has progressed considerably, but it is saturated on the other hand. 

How about the pixel structure if it is a form that is advantageous for recognition and 3D reconstruction. The theme on post-pixel may be discussed.

%
%
%
%

\hfill July 21, 2017

\ifCLASSOPTIONcaptionsoff
  \newpage
\fi

\end{document}